\documentclass{article}

\usepackage{arxiv}

\usepackage[utf8]{inputenc} 
\usepackage[T1]{fontenc}    
\usepackage{hyperref}       
\usepackage{url}            
\usepackage{booktabs}       
\usepackage{amsfonts}       
\usepackage{nicefrac}       
\usepackage{microtype}      
\usepackage{lipsum}		
\usepackage{graphicx}
\usepackage{natbib}
\usepackage{doi}

\usepackage{amsmath}

\newtheorem{theorem}{Theorem}

\title{Implicit Neural Representation of Tileable Material Textures}

\author{ 
Hallison Paz \\
	IMPA\\
	\texttt{hallpaz@impa.br} \\
	\And
 Tiago Novelo \\
	IMPA\\
	\texttt{tiago.novello@impa.br} \\
	\AND
	Luiz Velho \\
	IMPA \\
    lvelho@impa.br
}

\hypersetup{
pdftitle={A template for the arxiv style},
pdfsubject={cs.CV, q-bio.QM},
pdfauthor={Hallison Paz, Tiago Novello, Luiz Velho},
pdfkeywords={Implicit Neural Representation, Seamless Material, Sinusoidal Neural Networks},
}

\begin{document}
\maketitle

\begin{abstract}
	We explore sinusoidal neural networks to represent periodic tileable textures. 
Our approach leverages the Fourier series by initializing the first layer of a sinusoidal neural network with integer frequencies with a period $P$. 
We prove that the compositions of sinusoidal layers generate only integer frequencies with period $P$. As a result, our network learns a continuous representation of a periodic pattern, enabling direct evaluation at any spatial coordinate without the need for interpolation. 
\myhl{To enforce the resulting pattern to be tileable, we add a regularization term, based on the Poisson equation, to the loss function.}
Our proposed neural implicit representation is compact and enables efficient reconstruction of high-resolution textures with high visual fidelity and sharpness across multiple levels of detail. 
We present applications of our approach in the domain of anti-aliased surface.
\end{abstract}

\keywords{Implicit Neural Representation \and Seamless Material \and Sinusoidal Neural Networks}

\section{Introduction}

\textit{Textures} play a fundamental role in computer graphics, enabling the creation of visually appealing and realistic virtual environments. 
They model essential information of the surface materials in virtual scenes, enriching the visual experience with patterns that would be difficult to model otherwise. 
Within this scope, \textit{periodic textures} stand out as an important example that models repeating patterns. They are used to represent materials found in domains such as architecture, textiles, and industrial design. Capturing the essence of such textures is an important task to bridge the gap between virtual scenes and their real-world counterparts.

Traditionally, texture tiles are depicted as discrete digital images. Discrete representations introduce potential challenges in sampling and interpolation for operations. This is particularly important for achieving seamless transitions.

Recently, \textit{implicit neural representations} (INRs) have emerged as an alternative representation for images.
They are a class of neural networks that take the 2D coordinates as input and output the continuous colors. INRs are continuous, compact, fast to evaluate, and have high representation capacity.
Inspired by these properties, we propose to use INRs to represent seamless periodic textures. 
We focus on the subclass that uses the \textit{sine} as activation function --  \textit{sinusoidal INRs}~\cite{sitzmann2019siren}.

Training a sinusoidal INR to fit a given texture sample has the drawback of its image possibly being non-tileable.
To avoid this, we introduce a condition for an INR to be \textit{periodic} and propose a regularization term based on the \textit{Poisson equation}. 
Regarding the periodicity of an INR, we observe that if one takes the first layer of a {sinusoidal} INR and initializes it with integer frequencies with a period $P$, the entire INR will be periodic with period $P$. 
This is due to the fact that the composition of sinusoidal layers expands as a sum of sines with frequencies given by integer combinations of the input layer, and the result is similarly periodic~\cite{novello2022understanding, yuce2022structured}. 
We leverage this property to fit periodic signals such as image and texture representations.

The next step in our method involves defining a regularization term to \myhl{ensure that the image produced by a periodic INR forms a seamless, tileable texture -- meaning it should tile across the plane without displaying noticeable seams or discontinuities at the tile boundary}.
To achieve this, we employ the Poisson equation which enforces the INR gradient matches the original gradient within the domain (Poisson equation) and asks the INR to be equal to an image average on the domain border (periodic boundary values)~\cite{perez2023poisson}.
It is important to note that for this problem to be well-posed, the boundary must be closed. \myhl{However, since our network is periodic, we are solving the Poisson problem on the torus which allows us to also require gradient matching on the domain boundary (see the results in Section \ref{s:poisson-regularization} ).}

Besides the periodic representation, our method leads to better representation quality of images with fewer parameters in the architecture. Additionally, based on Fourier series, we design a frequency initialization that greatly decreases the amount of possible choices for frequencies. 
Our experiments demonstrate that this representation is suitable for representing periodic textures (and making them seamless) and can be easily integrated to the graphics pipeline for texture mapping.
In summary, our contributions are:
\begin{itemize}
\item We propose an initialization scheme for sinusoidal INRs that uses integer multiples of a period $P$ with reduced redundancy. We prove that this approach yields periodic INRs by demonstrating that the composition of periodic sinusoidal layers with period $P$ preserves periodicity.
\item \myhl{To ensure seamlessness at the boundary domain, we introduce a regularization term based on the Poisson equation.} 
We show that the resulting loss function can train a periodic INR to represent high-resolution textures with good quality and a reduced number of parameters. Moreover, these networks capture \textit{sharp} details in the data.
\item The proposed periodic INR framework showcases the ability to learn from specific segments of the ground-truth pattern without compromising information integrity.
\item We apply our method in texture mapping, showcasing its versatility across texture representation tasks, such as multiresolution~\cite{paz2023mr}.
\end{itemize}


\section{Background and related methods}

The scope of this work is to represent seamless material textures using \textit{implicit neural representations} (INRs). 
Thus, it aims to encode the material texture of a surface through a periodic neural network $f:\R^2\to \mathcal{C}$, where $\mathcal{C}$ is the \texttt{RGB} color space, that can be trained based on data samples. More generally, it could be extended to the other channels of a material definition.

In this sense, the context is texture creation methods (procedural-based, capture-based, AI-based, and manual) with the purpose of being applied to a 3D surface by a rendering system.

Generating high-quality textured materials is challenging as they need to be visually realistic, seamlessly tileable, and have a small impact on the memory consumption of the application. For this reason, these materials are often created manually by skilled artists.
In this work, we show that INRs can be used in those tasks.

Additionally, it is desirable to have an agnostic material representation based industry standards that is compatible with most rendering systems. One case is the Substance 3D Sampler from~\citet{substance_sampler}, a creation platform of material collections that includes many tools for that very purpose. 
Next, we discuss the most relevant works that are related to our method.

\subsection{Texture representations}

Texture is ubiquitous in computer graphics. 
It is the key to represent visual information for the synthesis of naturalistic textures and images from photographic exemplars.
Texture synthesis has grown into a mature field encompassing a variety of methods, such as: classic texture mapping \cite{blinn76}, procedural textures \cite{perlin-1985}, synthesis-by-example \cite{efros99}, and user-guided texture generation \cite{haeberli90}.

\citet{pauly-2009} gives an overview of example-based texture synthesis methods.
Also, 
\citet{etal-2010} provides an in-depth account of noise primitives for procedural texturing.
Recently, \citet{rethinkngtex} discusses alternative texture-mapping methods.

Most of the above methods target general texture patterns, however, there is a trend that investigates models of specific materials. This approach has the benefit of producing higher fidelity results by exploiting characteristics of important materials, such as wood.
\citet{dorsey-2004} studies aggregate materials that are common in nature and formed by small particles.

\subsection{Tileable Textures}
While there is a large body of research devoted to texture mapping, little work, however,  has been dedicated to synthesizing seamless tileable textures, which are important for the creation of materials.
These textures have the property that when laid out in a regular grid of tiles form a homogeneous appearance suitable for use in memory-sensitive real-time graphics.

\citet{tileinteractive} gives an overview of tile-based methods in computer graphics applications.
Additionally, we highlight the following methods.
\citet{tilehard} presents a tile-based texture mapping algorithm that
generates an arbitrarily large and non-periodic virtual texture map from the small set of stored texture tiles. 
\citet{Moritz2017Texture}
developed an approach to synthesize tileable textures using the PatchMatch texture synthesis algorithm.
These methods rely on a particular algorithm that must be incorporated into the rendering system. Our INR representation, once trained is exported to be evaluated on the graphics hardware or, alternatively, can be used to generate a material tile.

\citet{perez2023poisson} modeled the problem of making a rectangular domain texture tileable using a Poisson solver: the boundary conditions consist of forcing the opposite sides of the domain to be equals. In the interior, they asked for the gradient of the desired texture to be equal the ground-truth gradient.
This approach works only on pixel-based images.
Our method allows for the definition of a regularization term that compels the training of a periodic INR to produce a seamless, tileable texture. Furthermore, we can interchange the boundary and interior equations, given that our problem is defined on the 2D torus once the network is periodic.

\subsection{Neural Textures}
Texture representations strive to combine procedural methods with the fitting of data samples. \myhl{The main trade-off is between computation and memory}.
Principled solutions to such issue come from \emph{wavelets} and \emph{neural networks}: \citet{BAJAJ-2000} addresses the memory problem by encoding 3D textures using a wavelet-based method, \citet{Gutierrez-2019} describes a generative deep learning framework for 3D texture synthesis based on style transfer with results that are equivalent to patch based approaches.

Concerning tileable textures, \citet{deeptile} proposes an example-based texture synthesis using a deep learning process for creating tiles of arbitrary resolutions that resemble the structural components of an input texture.
\citet{zhou2022tilegen} developed a variant of StyleGAN whose architecture is modified to produce periodic material maps.
These works employ "data-based neural networks" and generative models, conversely, our method uses "coord-based neural networks" and a representational spectral model.

Finally, \citet{ntc2023}
developed a neural compression technique specifically designed for material textures. \citet{match}
introduced a deep neural feature-based graph creation method for constructing procedural materials.
In this respect, our~INR model is compact and can be incorporated into a neural rendering pipeline.

\subsection{Implicit neural representations}

Previous works such as \cite{park2019deepsdf, occupancy_networks, sitzmann2019siren, novello2022differential, 
novello2023neural,
silva2022mip-plicits,
yariv2020multiview} demonstrated how to represent and render surfaces as level sets of neural networks. 
Due to the spectral bias inherent in \textit{multilayer perceptrons} (MLPs) \cite{Rahaman2018O}, coordinate-based architectures often employ techniques like Fourier Feature Mapping \cite{tancik2020fourfeat} to map inputs to higher-dimensional spaces. Sinusoidal networks offer an alternative solution by utilizing periodic activation functions and operating on the spectral representation of the signal. However, the composition of sinusoidal layers can introduce new frequencies, making it challenging to train deep sinusoidal neural networks \cite{taming2017} and understand their behavior. 

For sinusoidal MLPs, \citet{sitzmann2019siren} proposed an initialization scheme for stable training and convergence. Additionally, \citet{paz2022} developed a multiresolution framework that controls the frequencies learned through initialization and data~filtering.

\textit{Multiplicative filter networks} (MFNs) \cite{fathony2020multiplicative} present an architecture that differs from MLPs by employing the Hadamard product instead of compositions. Building upon MFN, Bacon \cite{bacon2021} shows how to control frequencies in the network and represent signals at multiple scales. 
However, as each representational atom in the network spans fewer frequencies, a sinusoidal MFN may require more parameters to represent the same amount of frequencies. Additionally, Bacon truncates the frequency spectrum of the signal, leading to ringing artifacts.

We propose to use sinusoidal INRs to represent periodic textures. We prove that if the first layer is initialized with weights compatible with a period $P$, then the network is also periodic with period $P$. 

Regarding seamless textures, inspired by~\cite{perez2023poisson}, we model it as a term in the loss function using a Poisson problem. 
This strategy has been explored in INRs for tasks such as compositing gradients~\cite{sitzmann2019siren} and face morphing~\cite{schardong2023neural}. SIREN trains the INR using only the gradients, leading to the learning of the signal up to an integral constant, which is problematic when we have multiple channels. In contrast, our training scheme incorporates both gradient and image values.


\section{Periodic Neural Networks}
For simplicity, along this section, we assume the image's codomain to be $\R$, since the results can be easily extended to multiple channels.

A \textit{periodic image} $\gt{f}:\R^2\to \R$ is a function satisfying the property $\gt{f}(x) \!=\! \gt{f}(x + P)$ with $P=(P_1, P_2)\in \R^2$ giving the \textit{periods} along the axes $x_1$ and $x_2$.
For an example, consider the \textit{harmonic} $h(x_1, x_2)=a\sin(x_1\omega_1+x_2\omega_2+ \varphi)$ where $\omega_i\!=\!k_i\frac{2\pi}{P_i}$, with $k_i\in\Z$, are the \textit{frequencies}, $a$ is the \textit{amplitude}, and $\varphi$ is the \textit{phase shift}.
Such functions play a central role in \textit{Fourier series}, where the periodic function $\gt{f}$ can be approximated by a sum of harmonic functions.

\subsection{Shallow Sinusoidal networks}
Observe that summing harmonic functions also results in a \textit{sinusoidal network} $f:\R^2\to \R$ with a single layer:
\begin{align}\label{e-fourier_series}
    f(x) = L\circ s(x) =  c_0 + \sum_{i=1}^{n} c_i  \sin\Big(\dot{\omega_i}{ x}+ \varphi_i\Big)
\end{align}
where the \textit{first} layer $s:\R^2\to \R^{n}$ projects the input coordinates $x=(x_1,x_2)$ into a list of harmonics:
\begin{align}\label{e-firstlayer}
\displaystyle
    s(x)=\sin(\omega x +\varphi)=
    \left(
    \begin{array}{c}
        \sin\big(\dot{\omega_1}{x}+\varphi_1\big)\\
         {\footnotesize\vdots}\\
         \sin\big(\dot{\omega_{n}}{x}+\varphi_{n}\big)
    \end{array}
    \right).
\end{align}
Where $\dot{\omega_i}{x}=\omega_{i1}x_1+\omega_{i2}x_2$. Thus, the matrix $(\omega_1, \ldots, \omega_{n})$ gives the \textit{frequencies} and $(\varphi_1, \ldots, \varphi_{n})$ the \textit{phase shifts}.
The \textit{linear} layer $L$ combines the harmonics with the \textit{amplitudes} $c_i$ and adds a bias $c_0$.

For the network $f$ to be periodic with periods $(P_1,P_2)$ we can simply choose $\omega_{ij}=k_{ij}\frac{2\pi}{P_i}$, with $k_{ij}$ being integers.
Section~\ref{s-initialization} presents an initialization of $k_{ij}$ using ideas from Fourier series.

\subsection{Deep sinusoidal networks}\label{s-deep-networks}

We compose the first layer of $f$ with a hidden layer $h:\R^{n}\to \R^{m}$ to make $f$ deeper.
That is,
$f=L\circ h \circ \,s$, 
with $h\circ s(x):=\sin\big(W \cdot s(x)+b\big)$, where $W\in\R^{m\times n}$ is the \textit{hidden matrix}, and $b\in\R^{m}$ is the \textit{bias}.

\noindent The $i$th coordinate of $h$ one express as a \textit{sinusoidal perceptron}:
\begin{align*}
h_{i}\circ s(x)=\sin\left(\sum_{j=1}^{n} W_{ij} \sin\Big(\dot{\omega_j}{x}+\varphi_j\Big) + b_{i}\right).
\end{align*}
Thus, $f$ consists of multiple sine compositions. In a more general sense, a \textit{sinusoidal network} is a function constructed by combining sinusoidal perceptrons.
Furthermore, by adding more hidden layers to $f$, we obtain a sinusoidal \textit{multilayer perceptron} (MLP). Training such networks can be challenging because using the sine as an activation function may lead to instability in deep architectures~\cite{taming2017}. \citet{sitzmann2019siren} overcome this by proposing SIREN, which gives a special initialization scheme for sinusoidal MLPs, providing stability during training.
Clearly, the activation function is periodic, but the network $f$ may not.
Recall that even summing periodic function may result in non-periodic function: e.g. $\sin(x)+\sin\left(\sqrt{2}x\right)$. 
Furthermore, to accurately represent a periodic function, we must have control over the network's period.

In this work, we observe that by initializing the first layer $s$ of a sinusoidal MLP $f$ with integer frequencies, specifically $\omega_{ij}=k_{ij}\frac{2\pi}{P_i}$ where $k_{ij}\in\Z$, we can prove that $f$ will be periodic with period $P=(P_1,P_{2})$.
In other words, $f(x_1, x_2)=f(x_1+P_1, x_2+P_2)$.

For simplicity, let $f$ be a 1D function, i.e. $f:\R\to\R$, the general case follows analogously. 
Assuming the first layer to be periodic with period $P$, we see that its frequencies are expressed as $\omega_j=k_{j}\frac{2\pi}{P}$. 
Moreover, observe that if we prove that if each sinusoidal neuron $h(x)=\sin\left(\sum_{i=1}^n a_j\sin(\omega_j x + \varphi_i)\right)$ expands as a sum of harmonics with period $P$, then, the output of $h$ is periodic with period $P$. The proof of this fact follows from the following identity~\cite{novello2022understanding}:
\begin{align}\label{e-expansion}
    h(x)= \!\!\sum_{\textbf{l}\in\Z^n}\left[\prod_{i=1}^n J_{l_i}(a_i)\right]\sin\Big(\dot{\textbf{l}}{\omega x +\varphi}+ b\Big).
\end{align}
Since each frequency in this sum is written as $\dot{\textbf{l}}{\omega}=\frac{2\pi}{P}\dot{\textbf{l}}{\textbf{k}}$, we have that $h$ is periodic with period $P$.
We note that an expansion similar to Equation~\ref{e-expansion} was also presented in \cite{yuce2022structured}.

For sinusoidal MLP with two hidden layers, we truncate the expansion given by the expansion of the first layer. Thus the input of the second hidden layer is a finite sum of sines and Equation~\ref{e-expansion} can be applied.
For the truncation, we use the fact that the amplitudes
$\prod_{i=1}^n J_{l_i}(a_i)$ goes rapidly to zero as $\norm{\textbf{l}}_{\infty}$ increases. This is due to the following inequality~\cite{novello2022understanding}:
\begin{align}\label{e-upper-bound-freq}
    \abs{\prod_{i=1}^n J_{l_i}(a_i)}<
    \prod_{i=1}^n\frac{\left(\frac{\abs{a_i}}{2}\right)^{\norm{l_i}}}{\abs{l_i}!}.
\end{align}
Applying induction in the above procedure implies the result:
\begin{theorem}
\label{t-periodic}
    If the first layer of a sinusoidal MLP $f$ is periodic with period $P$, then $f$ is also periodic with period $P$.
\end{theorem}

Theorem~\ref{t-periodic} allows us to use sinusoidal MLPs to represent periodic functions. 
We define a \textit{periodic INR} to be a sinusoidal MLP such that its first layer is periodic. 

\subsection{Multiresolution sinusoidal networks}\label{s-mr-networks}
\label{s-multiresolution}
To represent textures using INRs we need to represent them in multiresolution. For this, we adopt the \textit{multiresolution network} (MRnet)~\cite{paz2022} as a representation. Here, a MRnet is a network $f\!:\!\R^2\!\times\! [0,N]\!\to\! \R$ defined as a sum of $N$ periodic INR $g_i:\R^2\to\R$:
\begin{align}\label{e-mrnet}
f(x,t) = c_0(t)g_0(x) + \cdots + c_{N-1}(t)g_{N-1}(x),
\end{align}
The contribution of each \textit{stage} $g_i$ in Equation~\ref{e-mrnet} is controlled by $
c_i(t)\!=\!\max\big\{0, \min\big\{1, t-i\big\}\big\}.$
This allows us to navigate in the multiresolution using a parameter $t\!=\!k+\delta$ with $k\in\mathbb{N}$ and $0\leq\delta\leq 1$:
\begin{align}\label{e-mrnet2}
f(x,t)=g_0(x)+\dots + g_k(x)+\delta g_{k+1}(x).
\end{align}

The \textit{level of details} $f(\cdot, t)$ evolve continuously.
We set the first layer of each periodic INR $g_i$ with integer frequencies, since Theorem~\ref{t-periodic} says that in this case $f$ will \textit{periodic} with respect to $x$.

\subsection{Frequency initialization}
\label{s-initialization}
The first layer of a sinusoidal INR projects the input coordinates into a list of sines (Eq.~\ref{e-firstlayer}). Next, we show that in a shallow network, this gives the frequencies of the signal represented by a Fourier series. Thus, the initialization of the first layer is important for network performance.
Following the conclusions of Theorem~\ref{t-periodic}, we define the first layer's frequencies to be integer multiples of a period.

Let $\gt{f}\!:\!\R^2\!\to\! \R$ be a periodic image (the \textit{ground-truth}) with period~$P$ and ${f}\!:\!\R^2\!\to\! \R$ be a periodic INR.
To define the integer matrix $k_{jl}$ of the first layer of $f$ we follow ideas from Fourier~series.
If $\gt{f}\in L^1(P)$, the \textit{Fourier theorem} says that it expands in a series:
\begin{align}\label{e-fourier-expansion}
    \gt{f}(x)=\sum_{\textbf{k}\in \Z^2} c_\textbf{k} \text{e}^{i\dot{\omega_{{\textbf{k}}}}{ x}},
\end{align}
where $c_\textbf{k}$ are the \textit{Fourier coefficients}, and $\omega_{\textbf{k}}=2\pi\left(\frac{k_1}{P_1}, \frac{k_2}{P_2}\right)$ with $\textbf{k}=(k_1,k_2)$. 
In practice, we truncate the Fourier series summing over the integers $\textbf{k}\in[-B, B]^2$, where $B$ is a \textit{bandlimit}.

On the other hand, assuming $f$ to be a network with a single layer (Eq~\eqref{e-fourier_series}) and adding $\frac{\pi}{2}$ to the biases, we obtain:
\begin{align}\small
    f(x) &=  \frac{a_0}{2} + \sum_{k=1}^{n} a_k  \cos\big(\dot{\omega_k}{ x}+ \varphi_k\big)\\
    &=  c_0 + \sum_{k=1}^{n} c_k  \text{e}^{i\dot{\omega_{{k}}}{ x}}+ \sum_{k=-n}^{-1} c_k  \text{e}^{i\dot{-\omega_{\abs{k}}}{ x}},\label{e-mlp2complex}
\end{align}
where $c_k=\frac{a_{\abs{k}}}{2}\text{e}^{\text{sign}(k)i\varphi_{\abs{k}}}$ with $\varphi_0=0$, and $\omega_{j}=2\pi\left(\frac{k_{j,1}}{P_1}, \frac{k_{j,2}}{P_2}\right)$. 
%
Thus, for $f$ to approximate the truncated series in Eq.~\eqref{e-fourier-expansion}, we can set
the integers $k_{j}$ in a set $\textbf{K}\subset [-B,B]^n$ such that $-\textbf{K} = [-B,B]^n\setminus \textbf{K}$.

When $f$ contains hidden layers, we initialize its parameters following the scheme in ~\cite{sitzmann2019siren}. 
Notice that we initialized $f$ with a finite number of frequencies in the frequency set $\textbf{K}$. However, 
the layer composition produces many other frequencies as the depth of the network increases.
Equation~\ref{e-expansion} justifies this claim.

Regarding the frequency initialization of each stage $g_i$, we split the frequency set $\textbf{K}$ in $N$ subsets $\textbf{K}_i$ sorted by length.
Thus, the first layer of $g_i$ is initialized with the frequencies in $\textbf{K}_i$.
This initializes subsequent stages using only frequencies that were not chosen in previous stages enforcing the early stages to contain the lower frequencies, then as stages advance we add the higher frequencies.

\subsection{Network training}
\label{s-training}
Let $\gt{f}:\Omega\subset\R^2\to \mathcal{C}$ be an image (\textit{ground-truth}) defined in the rectangular domain $\Omega=[0,P_1]\times[0,P_2]$. 
We aim to approximate $\gt{f}$ by a periodic INR ${f}:\R^2\to \mathcal{C}$ with periods $(P_1, P_2)$ such that the resulting image is seamless. 
For this, define the weights $\omega_{ij}$ of the first layer of $f$ in the form $k_{ij}\frac{2\pi}{P_i}$ with $k_{ij}\in \Z^2$.
Again, this implies that the first layer is periodic with periods $(P_1,P_2)$, then, Theorem~\ref{t-periodic} implies that $f$ is also periodic with the same periods. 
To force the resulting texture to be seamless at the boundary $\partial \Omega$ of $\Omega$, we design a loss function based on a \textit{Poisson problem}.

Specifically, we use the Jacobian $\jac{\gt{f}}$ of $\gt{f}$ to train the periodic INR $f$.
For this, we define a matrix field $U$ such that its primitive approximates a seamless tileable texture at $\partial \Omega$.
Then we enforce $\gt{f}=f$ at some region of $\Omega$. This can be modeled using:
\begin{align}\label{e-gradient-interpolant}\small
\min \int_{\Omega} \lambda\norm{{\jac{f}-U}}^2dx \text{ subject to } (1-\lambda)(\gt{f}-f)=0 \text{ in } \Omega.
\end{align}
Where $\lambda:\Omega\to [0,1]$ is a weight function indicating that $U$ should match the Jacobian of $f$ if it is close to one ($\lambda\approx 1$), and enforces $\gt{f}=f$, otherwise. 
 We propose to use this variational problem to define the following loss function to train the parameters $\theta$ of $f$.
\begin{align}\label{e-blending-no-grad}\small
\mathscr{L}(\theta)={\int_{\Omega} \lambda\norm{{\jac{f}-U}}^2dx} + {\int_{\Omega} (1-\lambda)\big(\gt{f}-f\big)^2dx}.
\end{align}
\noindent
Thus, $\mathscr{L}$ trains $f$ to \textit{seamless clone} the primitive of $U$ to $f$ in $\Omega$.
Unlike classical approaches that rely on pixel manipulation, seamless cloning operates on the image gradients.
Note that $\lambda$ depends on the position $x$. In practice, we consider it to have high values near $\partial \Omega$. As a result, the training forces the matching of the Jacobian of $f$ with $U$ near $\partial \Omega$. In fact, since $f$ is periodic, we are training it on the torus given by the identification of the opposite edges of $\Omega$.
Classical methods do not provide such flexibility.


\section{Experiments}

This section presents experiments using our method for tileable material texture representation. We begin by fitting a tileable pattern across multiple scales, and assess its performance qualitatively. 
Then, we train the network using a sample containing repetitions of the pattern, with a fundamental period smaller than the training domain. For this, we utilize masks to exclude certain parts of the image during training and examine the network's ability to reconstruct them. 
Additionally, we address non-tileable patterns that can be made seamless using Poisson regularization. Finally, we provide comparative evaluations with related methods.

\subsection{Seamless Tileable Materials}\label{s-multires-2d}

We start with a single tile, matching the fundamental period of a texture, with a resolution of $1024^2$ pixels and $8$ bits per color channel in \texttt{RGB} space. We convert the image to YCbCr and train a MRnet, using the framework in \cite{paz2022}, consisting of a $6$ periodic INRs (stages).
We use periodic INRs with a single hidden layer, and heterogeneous width, as experiments indicated that we need fewer parameters to fit an image of lower resolution and low-frequency details. The initialization uses the scheme described in Sec~\ref{s-initialization}. The architecture of the hidden layer (number of input/output neurons) and the band limits of the first layer are in Table \ref{tab:mnet-architecture}.

\begin{table}[h]
\centering
\small
\begin{tabular}{|l|c|l|l|}
\hline
\textbf{Stage} & \multicolumn{1}{l|}{\textbf{Band-Limit}} & \multicolumn{1}{l|}{\textbf{Hidden Input}} & \multicolumn{1}{l|}{\textbf{Hidden Output}} \\ \hline
 0     & $[0, 3]\times[-3, 3]$                       & 24                                & 32                                 \\
 1     & $[0, 6]\times[-6, 6]$                       & 48                                & 32                                 \\
 2     & $[0, 12]\times[-12, 12]$                     & 80                                & 64                                 \\
 3     & $[0, 24]\times[-24, 24]$                     & 192                               & 160                                \\
 4     & $[0, 56]\times[-56, 56]$                     & 384                               & 256                                \\
 5     & $[0, 128]\times[-128, 128]$                   & 1024                              & 512                                \\ \hline
\end{tabular}
\caption{M-Net architecture.}
\end{table}\label{tab:mnet-architecture}

\begin{figure}[h]
\centering
\includegraphics[width=0.38\linewidth]{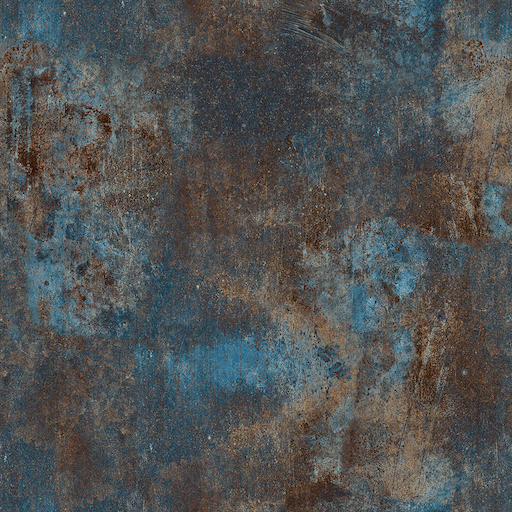}
\includegraphics[width=0.38\linewidth]{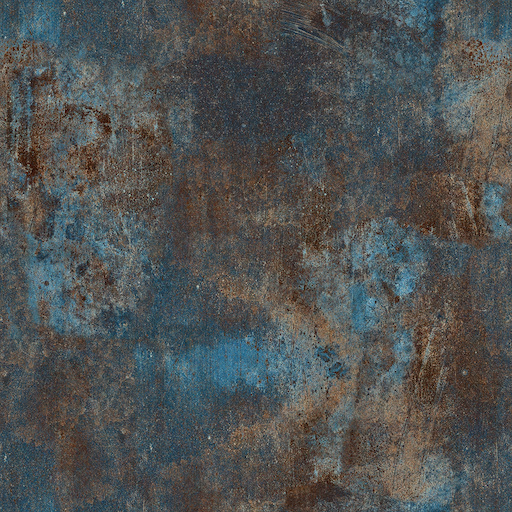}
\caption{Original image (left); reconstruction of the network (right) 
}
\label{f:rec_gt}
\end{figure}

Figure \ref{f:rec_gt} presents a qualitative comparison between the original image and the reconstructed image using our method. 
Observe that the images exhibit a high degree of similarity with a PSNR of 31.8 dB.
Additionally, our model achieves this level of fidelity using 855,572 parameters, which is less than the number of pixels in the image (1,048,576). Our model also demonstrated the ability to encode information at multiresolution giving an even better compression compared to a traditional Mipmap.

\begin{figure}[!h]
\centering
\includegraphics[width=0.76\linewidth]{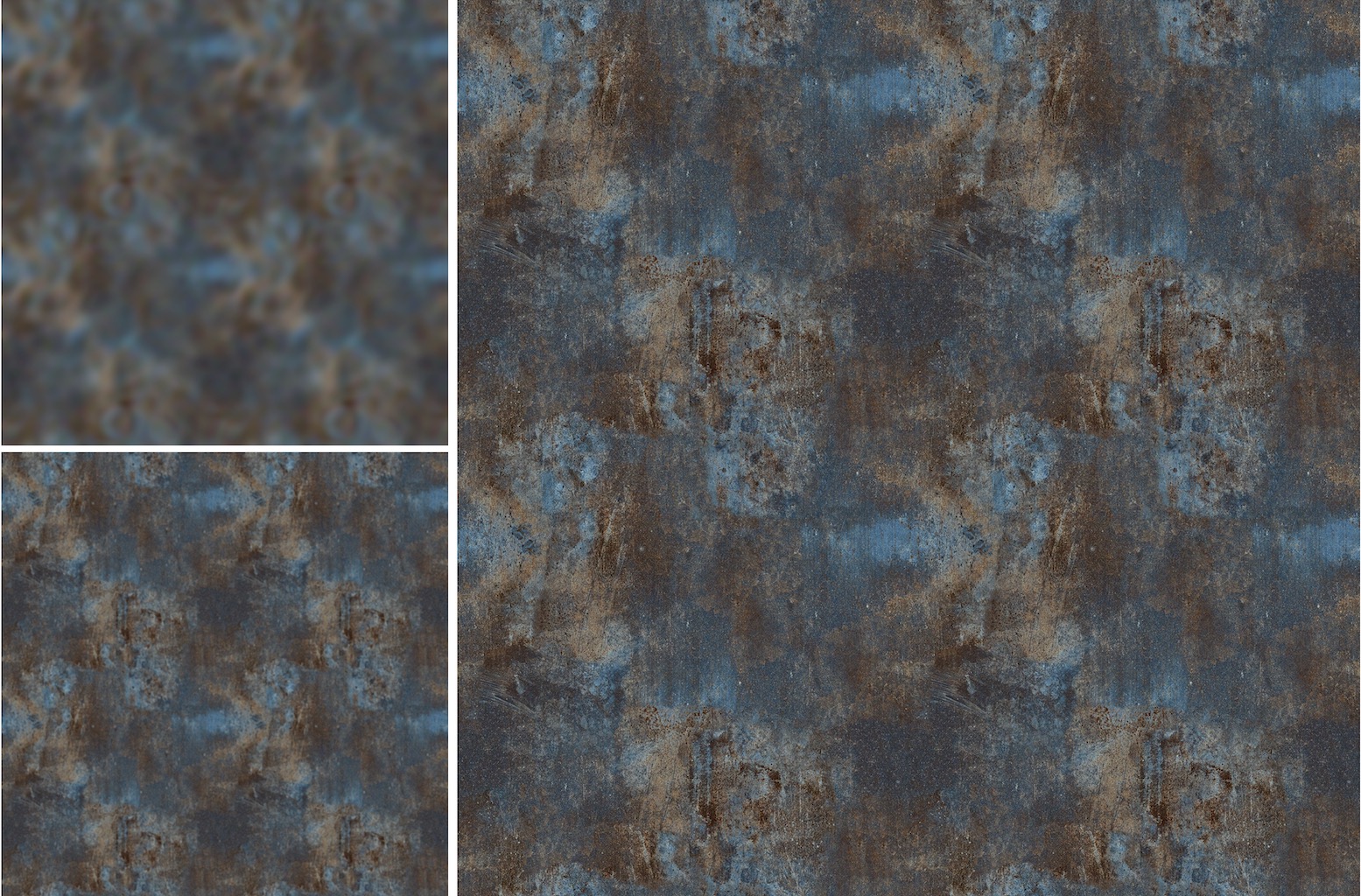}
\caption{Reconstructed multiresolution levels extrapolation. Top left: level 2; bottom left: level 4; right: level 6.}
\label{f:mr-periodic}
\end{figure}

To verify the periodicity of our INR model (Secs \ref{s-deep-networks} and \ref{s-mr-networks}), we evaluate the network in a larger domain ($[-2, 2]^2$) than its training domain $[-1, 1]^2$; Figure~\ref{f:mr-periodic} illustrates it for levels of detail 2, 4, and 6.

\subsection{Repeating Patterns in Sample}

We can also represent a texture with a fundamental period smaller than the sample size. 
This means that the periodic pattern repeats multiple times within the sample. 
To address this, we can specify the fundamental period either by prior knowledge or through pre-processing of the image, and train the network using the entire data or only a specific portion of it.

\begin{figure}[!h]
\centering
\includegraphics[width=0.24\linewidth]{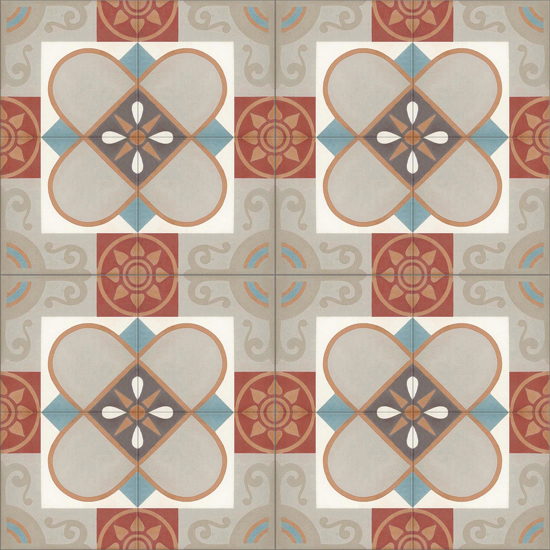}
\includegraphics[width=0.24\linewidth]{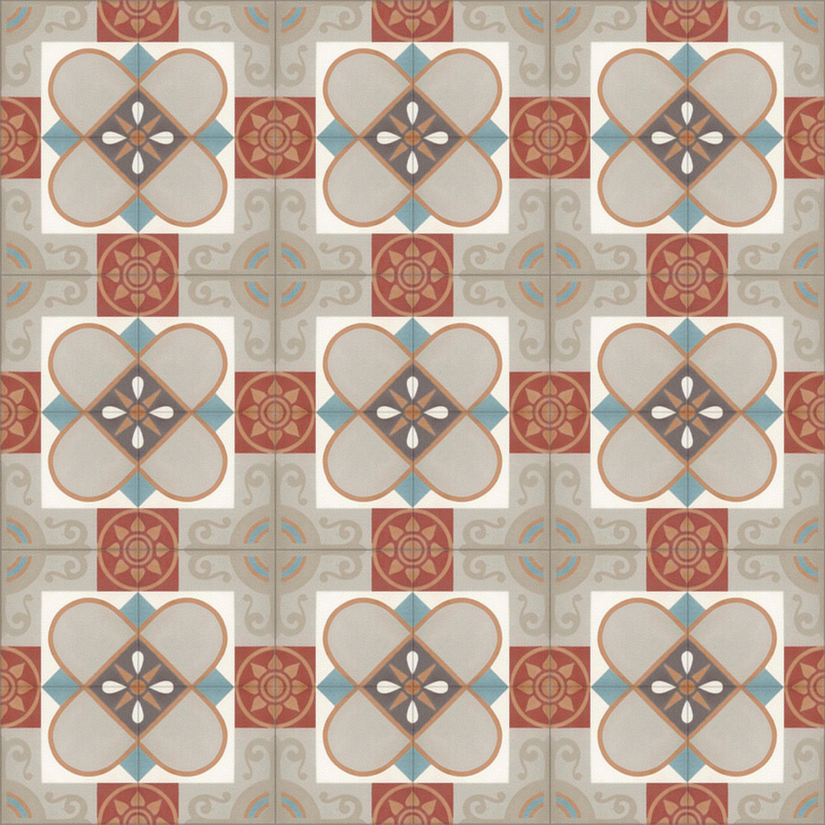}
\includegraphics[width=0.24\linewidth]{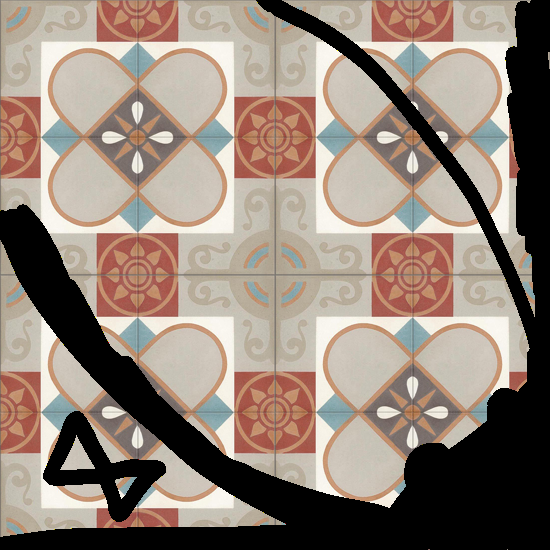}
\includegraphics[width=0.24\linewidth]{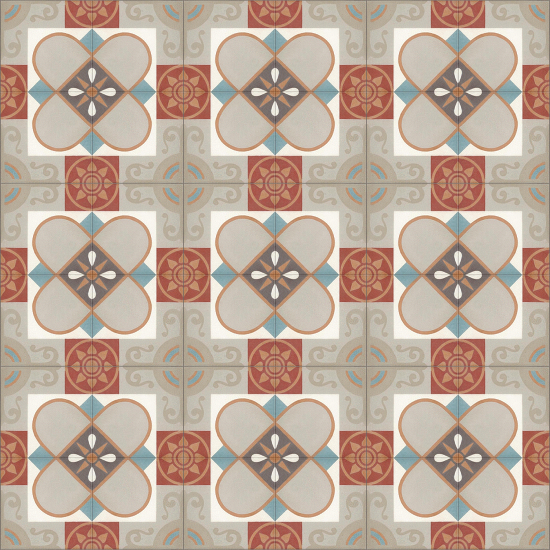}

\centerline{(a)\hfil\hfil(b)\hfil\hfil(c)\hfil\hfil(d)}

\caption{(a) Full training data. (b) Network reconstruction from full data. (c) Masked training data; black pixels were not provided in training. (d) Network reconstruction from masked data.}
\label{f:mask_nomask}
\end{figure}

Figure \ref{f:mask_nomask} showcases a texture sample where the pattern is repeated twice, and its extrapolated reconstructions in the region $[0, 3]^2$. The training considered $[-1, 1]^2$, with a specified period of 1. 
Figures~\ref{f:mask_nomask} (a) and (c) show the training data, while the network reconstruction is given by (b) and (d), respectively. 
In (a-b), all coordinates within the region are provided, thus the network learns a periodic representation of the pattern. 
In (c-d), we employ a mask to exclude a small part while preserving a contiguous region that contains the fundamental period. Again, our method reconstructs the periodic pattern across the entire region. This demonstrates that the over-determined problem does not negatively impact the results. 

\begin{figure}[!h]
\centering
\includegraphics[width=0.27\linewidth]{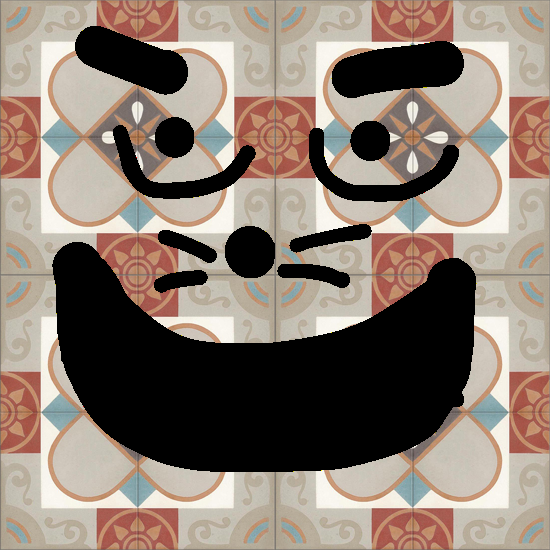}
\includegraphics[width=0.27\linewidth]{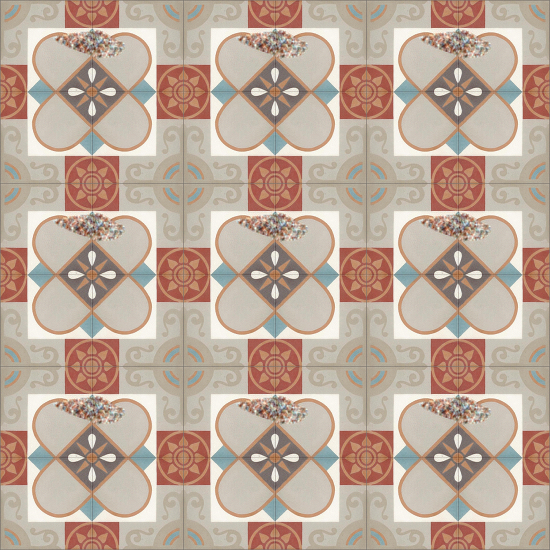}
\includegraphics[width=0.27\linewidth]{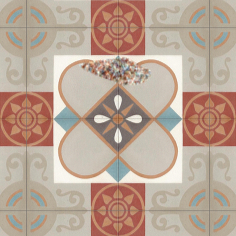}
\caption{From left to right: training data, network reconstruction in $[0, 3]^2$, and zoom in one tile to highlight the reconstruction artifact.}
\label{f:artifact}
\end{figure}

In the next experiment, a few points have no correspondents visible in any of the periods. This generates an artifact that is noticed across all occurrences of that pattern (Figure \ref{f:artifact}).

\begin{figure}[!h]
\centering
\includegraphics[width=0.4\linewidth]{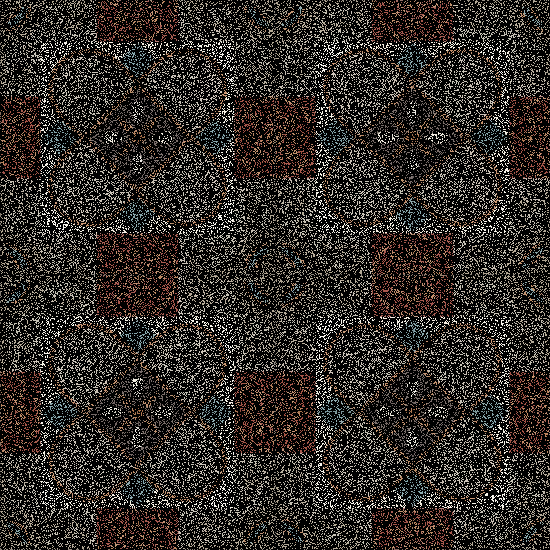}
\includegraphics[width=0.4\linewidth]{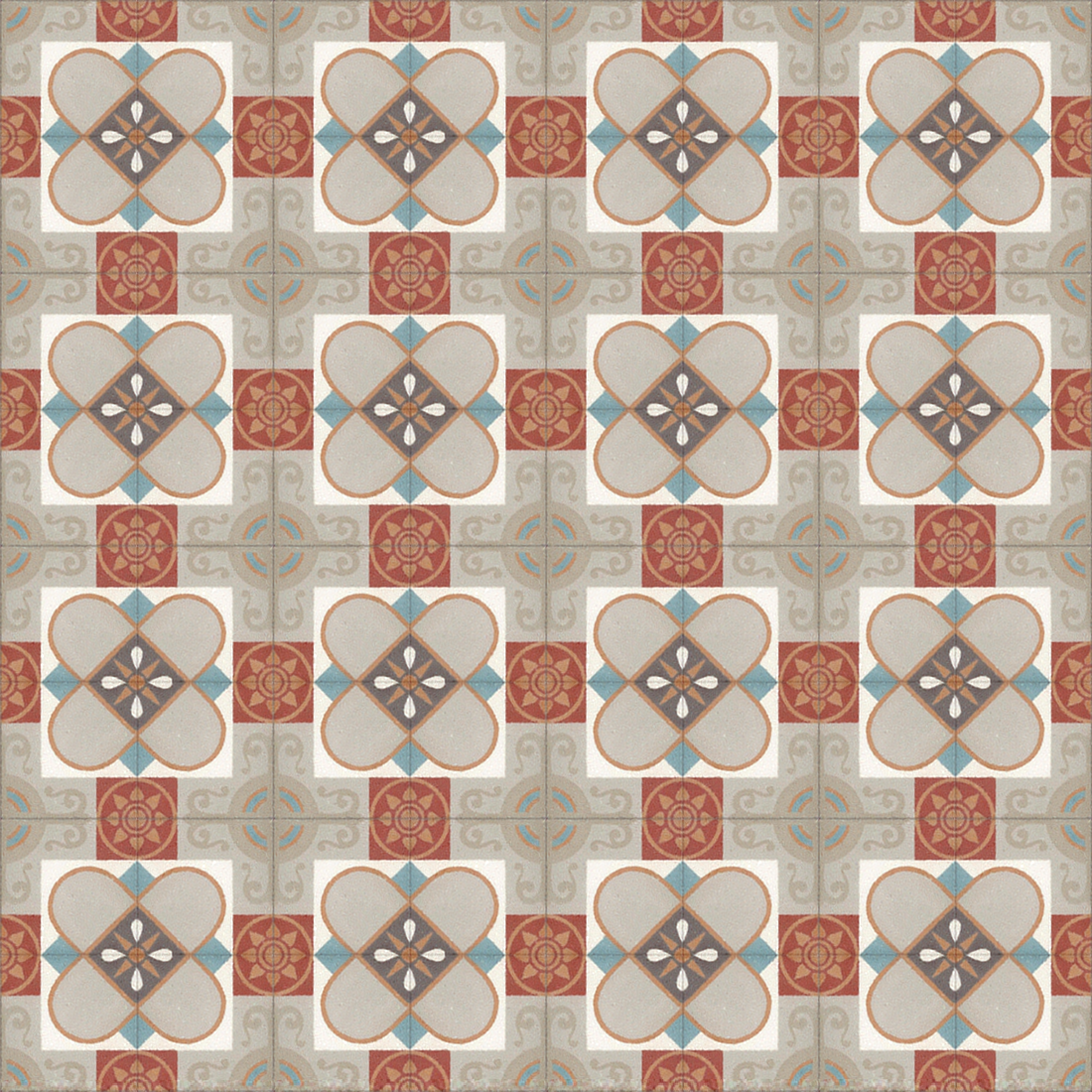}

\caption{On the left, the masked input data, where the black pixels were removed from training. On the right, the network reconstruction.~\\
}
\label{f:masked_recontruction}
\end{figure}

Next, we selectively remove points from the training data encouraging a sparse representation of the pattern. We generate the mask by randomly selecting a pixel and subsequently discarding all pixels that correspond to the selected one based on the pattern's period. 
Consequently, the mask encompasses exactly $1/4$ of the total pixels. Figure \ref{f:masked_recontruction} shows the training data, where the black pixels were removed from training, and the learned texture in $[2, 6]^2$. 

Even in this case, our method captures the underlying periodicity. This provides evidence that we can use the training data at any part of the domain. In other words, we only need a single sample from the class of equivalence defined by the periodicity to reconstruct the image without requiring any additional regularization.

\subsection{Seamless with Poisson Regularization}\label{s:poisson-regularization}

When obtaining a sample for a material, the inherent irregularity in the sample's pattern poses a challenge in seamlessly stitching it together. Achieving a seamless material is crucial for numerous applications. This section details the application of our methodology to generate seamless results from material patches.

We train a periodic INR for this task using the loss function of Section~\ref{s-training} which is based on the Poisson equation. In regard to classical Poisson problem solutions, a key observation of our method is the inversion of boundary and interior conditions. 
We employ masks (the function $\lambda$ in Eq~\ref{e-blending-no-grad}) to delineate the supervision of gradients at the border, while color values are supervised within the interior of the region. Given that our work is situated in the realm of periodic functions, the problem domain is equivalently represented as a torus, thus obviating traditional boundaries.

\begin{figure}[!h]
\centering
\includegraphics[width=0.90\linewidth]{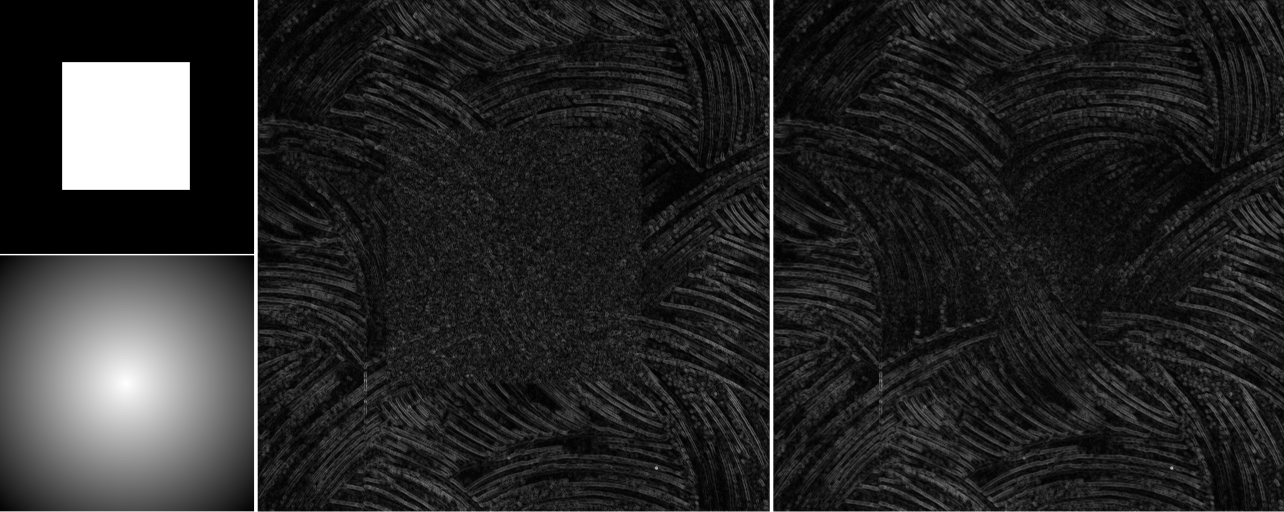}
\caption{Binary mask (top), soft mask (bottom), and the normalized magnitude of the gradients of the trained networks. This is the gradient of the texture in Line 2 of Fig \ref{f:seamless_examples}.}
\label{f:training_masks}
\end{figure}

While binary masks can produce satisfactory outcomes, they often introduce artifacts at the gradient of the reconstructed network (Figure \ref{f:training_masks}), requiring customized adjusts, per image, of the weights of the loss function components. Consequently, we opted for soft masks computed via a distance function to the center of the mask. 
In our experiments, we utilized the $L_2$ distance, but any $L_p$ distance can be employed as a parameter for refining the results. 

\begin{figure}[!h]
\centering
\includegraphics[width=0.30\linewidth]{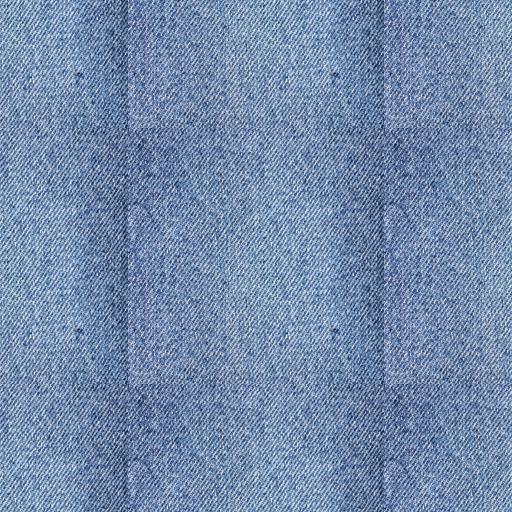}
\includegraphics[width=0.30\linewidth]{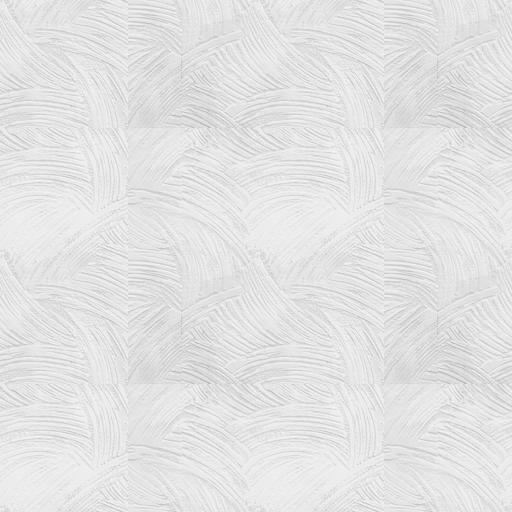}
\includegraphics[width=0.30\linewidth]{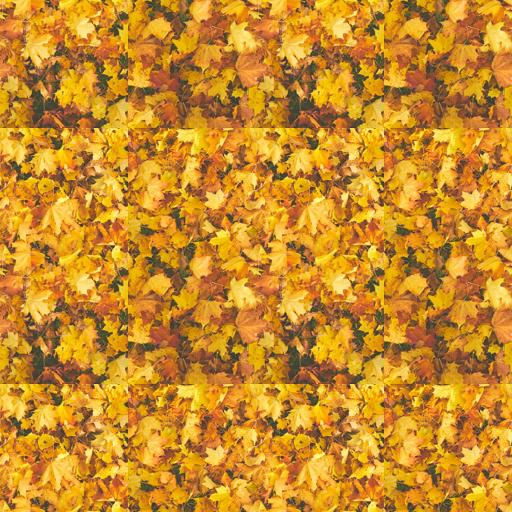}

\includegraphics[width=0.30\linewidth]{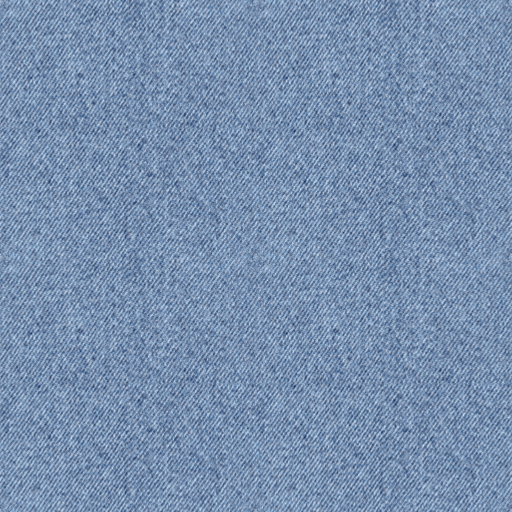}
\includegraphics[width=0.30\linewidth]{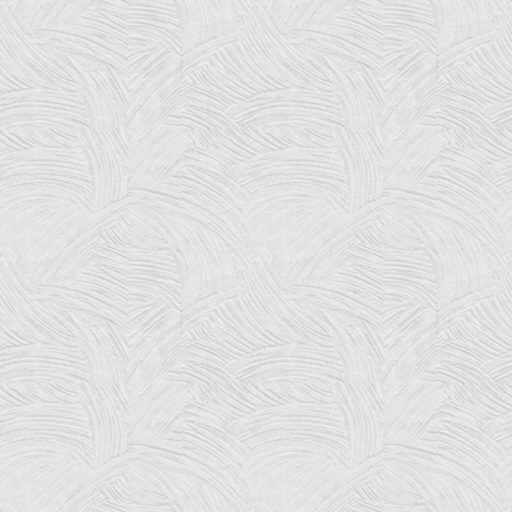}
\includegraphics[width=0.30\linewidth]{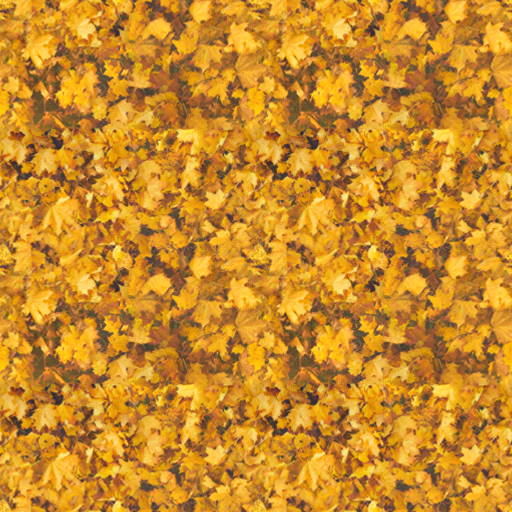}
\caption{Seamless reconstruction of jeans fabric (only fine details) on the left. Seamless reconstruction of a patch of a wall (medium details) on the middle; note that although this texture looks homogeneous it has fine details as can be seen on its gradient in Fig ~\ref{f:training_masks}. Finally, the third column gives the reconstruction of the leaves textures containing big details.
}
\label{f:seamless_examples}
\end{figure}

Figure~\ref{f:seamless_examples} illustrates several seamless reconstructions (second column) of non-tileable material samples (first column).
The first line shows an example of a fabric patch containing only fine details.
The second line gives the case of a pattern featuring medium geometric details. Meanwhile, the third line presents the reconstruction of a pattern characterized by larger details. 
Note that in all instances, our method adeptly conceals seams through the uniformization of photometric discrepancies. 
This results in a visually cohesive representation that effectively camouflages the transitions between material patches.

\subsection{Comparisons}

\begin{figure}[!h]
\centering
\includegraphics[width=0.36\linewidth]{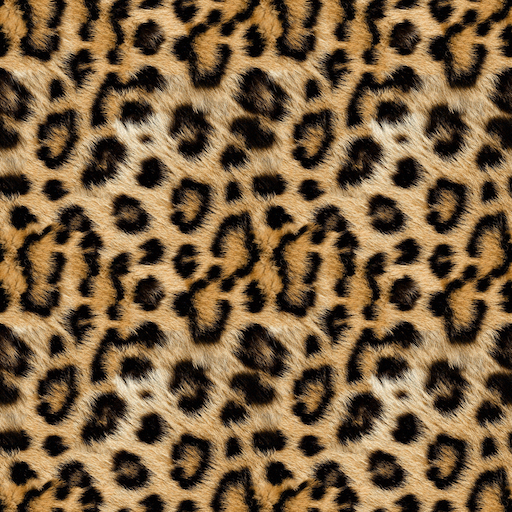}
\includegraphics[width=0.36\linewidth]{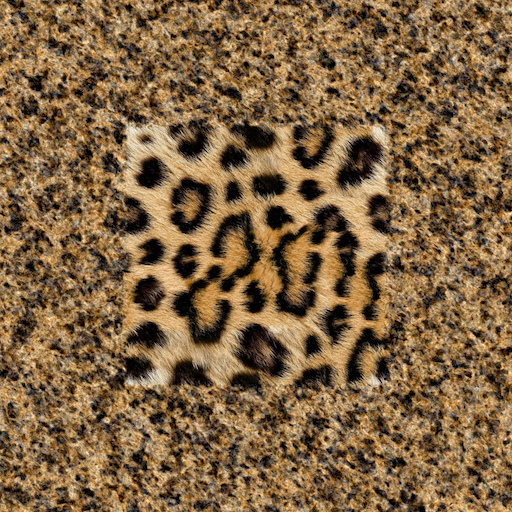}
\caption{Comparing our method (left) with Siren (right) on extrapolation.}
\label{f:comparison_siren}
\end{figure}

While SIREN and MRnet employ periodic activation functions, this is insufficient for learning periodic images with supervision in a restricted domain. Figure \ref{f:comparison_siren} shows a comparison between image reconstruction/extrapolation using SIREN and our periodic model. 

\noindent For this, we use an identical architecture to SIREN: periodic INR with 3 hidden layers, of the form $\R^{256}\to \R^{256}$. 
The key difference lies in the initialization: we randomly selected 256 pairs of integer frequencies from the range $[0, 30]\times[-30, 30]$, whereas SIREN follows the initialization given in \cite{sitzmann2019siren} with $\omega_0=30$. The images were trained on a $512\times512$ grid within the region $[-1, 1]$ for 500 epochs. The extrapolations are displayed in $[-2, 2]$.

It is worth noting that beyond the domain of supervision, the SIREN's reconstruction is very noisy, while our approach captures the underlying periodic pattern. 

\begin{figure}[!h]
\centering
\includegraphics[width=0.36\linewidth]{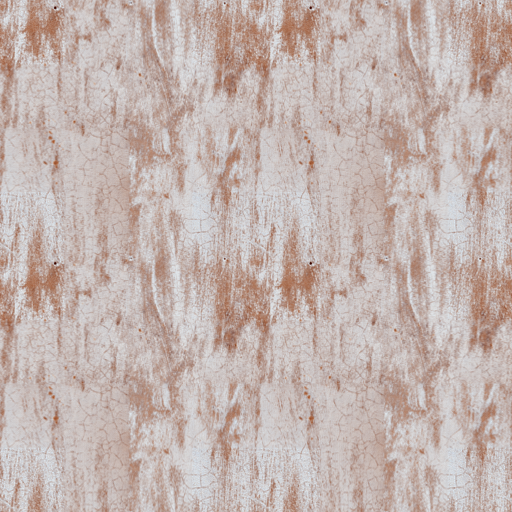}
\includegraphics[width=0.36\linewidth]{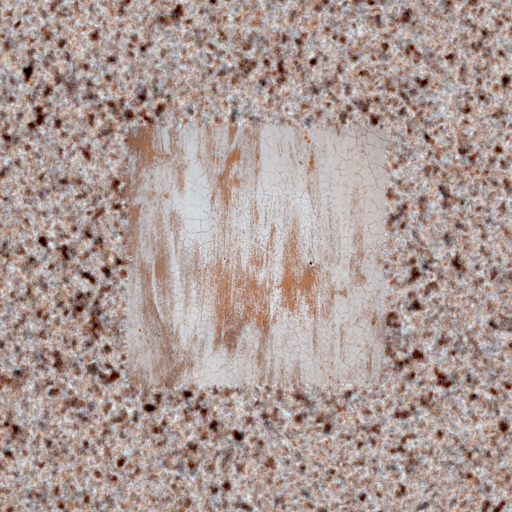}
\caption{Non-tileable patten. Our periodic initialization vs Siren's.}
\label{f:comparison_siren_nontileable}
\end{figure}

We also applied the Poisson regularization, using binary masks, to reconstruct non-tileable patterns in a seamless way using both SIREN and our periodic initialization, see Figure \ref{f:comparison_siren_nontileable}. Again, SIREN's reconstruction displays only noise outside the supervised domain, while our periodic initialization reconstructs a seamless texture.

\begin{figure}[!h]
\centering
\includegraphics[width=0.36\linewidth]{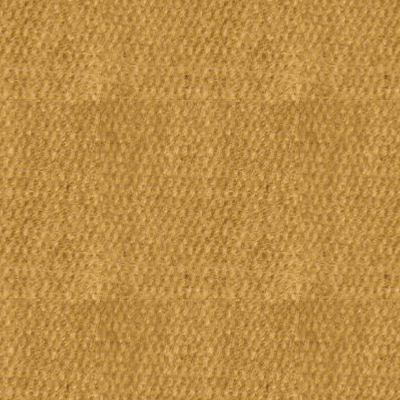}
\includegraphics[width=0.36\linewidth]{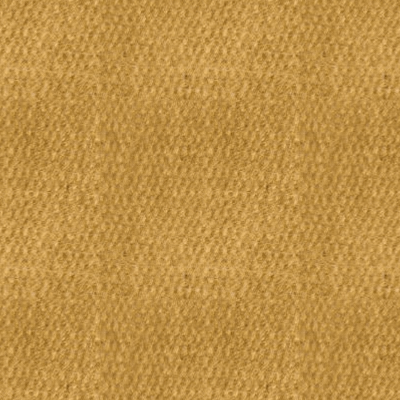}
\caption{Average border (left). Gradient Border (right).}
\label{f:average_border}
\end{figure}

To reconstruct seamless tiles using Poisson equation, \citet{perez2023poisson} suggests to manipulate the pixels intensities at the borders so that left = right and top = bottom, which can be achieved by averaging these values. These intensities should be given as boundary conditions while the gradients should be given at all other positions. In Figure \ref{f:average_border},  we compare this strategy with the one presented in Section \ref{s:poisson-regularization}. Note that our approach gives a more uniform result, better reducing the photometric differences present in the pattern.

\section{Applications}\label{s-applications}

\citet{paz2022} have demonstrated MRNet capabilities of encoding mip-mapping and displaying an anti-aliased version of an image, adapted to the required resolution. This section demonstrates practical applications of this property in mapping textures to surfaces.

\subsection{Surface Texture Mapping}

\begin{figure}[!h]
\centering
\includegraphics[width=0.20\linewidth]{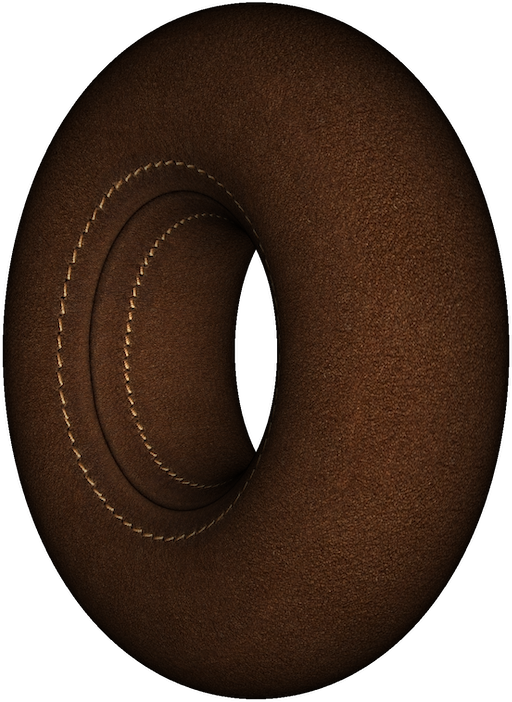}
\includegraphics[width=0.20\linewidth]{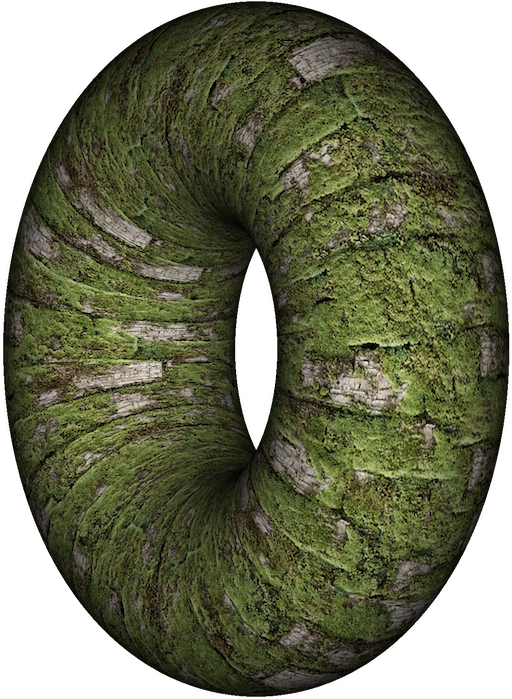}
\includegraphics[width=0.20\linewidth]{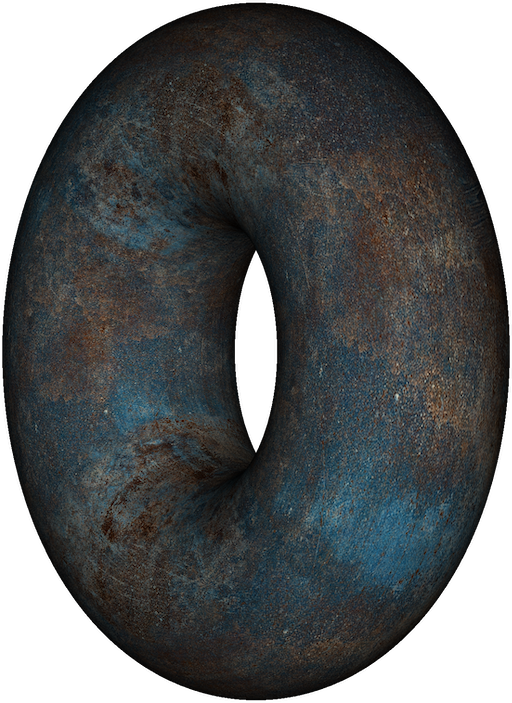}
\includegraphics[width=0.20\linewidth]{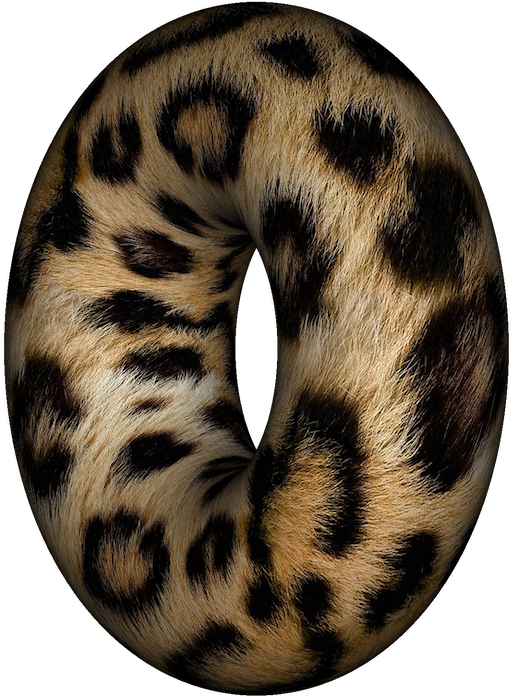}
\includegraphics[width=0.20\linewidth]{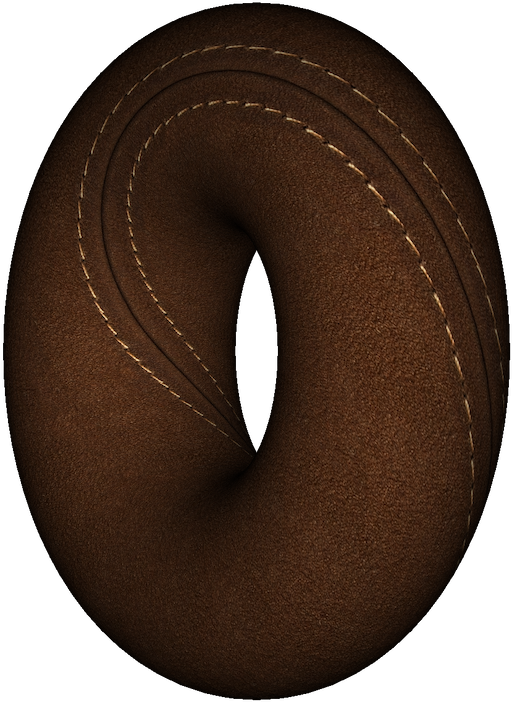}
\includegraphics[width=0.20\linewidth]{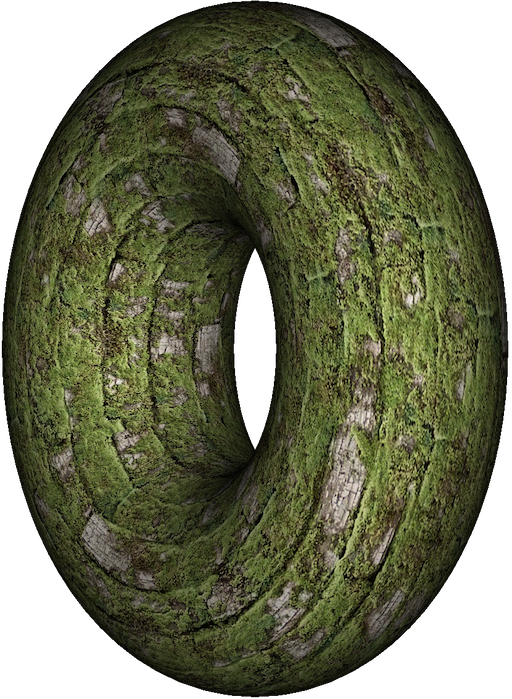}
\includegraphics[width=0.20\linewidth]{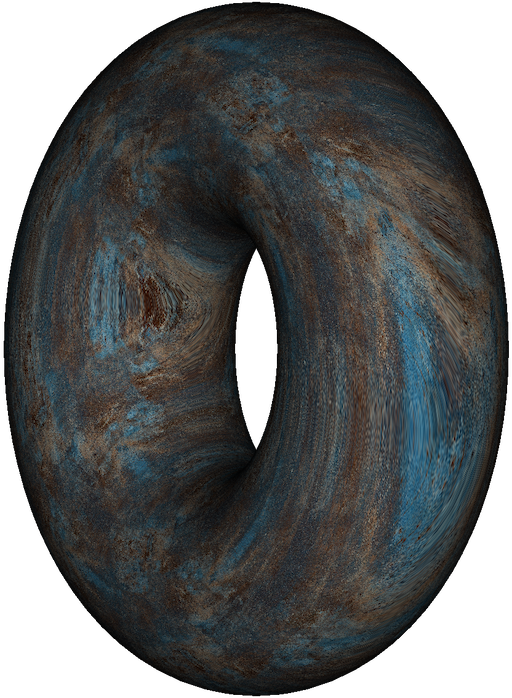}
\includegraphics[width=0.20\linewidth]{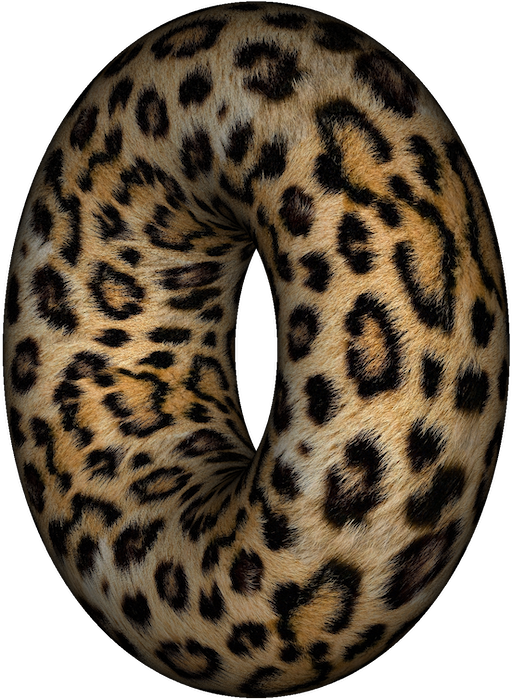}
\caption{Neural texture mapping on the torus using its $uv$-coordinates. Line 1 shows four texture examples. Line 2 presents these textures affected by different torus transformations.}
\label{f:surface_texture_mapping}
\end{figure}

We present a simple renderer to show the application of our networks in surface texture mapping (Fig.\ref{f:surface_texture_mapping}). We directly evaluate the network at the $uv$-coordinates of each fragment in a rendering pipeline. 
First, we map the $uv$-coordinates from the range $[0, 1]$ to  $[-1, 1]$; then, for the torus, we apply a factor of 2 in the $u$ coordinate for a better aspect to the rendering.
We can also scale the coordinates so that it spans multiple periods of the texture. 

Note that as our network is continuous in space and encodes levels of details, we can evaluate it in any coordinates, and interpolate between levels, akin to Mip-Map \cite{mipmap83}. 
In this regard, we can perform anti-aliasing similarly to the results in \cite{paz2023mr}.

\section{Conclusion and Future Work}

We presented an unified framework for modeling seamless textures by incorporating Fourier series ideas into neural network architectures. 
Our approach demonstrates the ability to train the network using partial texture data, making it more compact, and capturing sharper details compared to the ground truth in certain cases.

While we have made significant progress in understanding the impact of sinusoidal network initialization on representation capacity, there are still paths for future exploration. Although we have constrained the space of frequency choices by selecting integer multiples of a period and partitioning the frequency space, determining the appropriate band limit remains an empirical task. In our current approach, we freeze the weights of the first layer to maintain their constancy throughout training. However, finding a way to make these weights learnable parameters while preserving the desired period is a valuable research direction.

Extending our approach to higher dimensions is a direction for future work. Additionally, exploring the application of our method to represent other graphical objects, such as hypertextures, \citet{hypertexture}, by leveraging the versatility of vector fields in combination with surface representations, holds great potential.

To fully integrate neural networks as primitives in a graphics pipeline, it is crucial to develop methods for operating and editing them. The functional structure of sinusoidal INRs provides opportunities to explore algebraic structures for network manipulation. 

Periodic INRs have high-capacity representation. When restricted to periodic functions domain, they can naturally compress the images, specially in multiresolution. However, investigating compression techniques tailored for sinusoidal networks is essential for achieving compact storage and efficient transmission over computer networks. Exploring the distribution of signal frequencies to inform network initialization is a promising direction in this regard.

\bibliographystyle{unsrtnat}
\bibliography{references}  

\end{document}